# Applicability and Surrogacy of Uncorrelated Airspace Encounter Models at Low Altitudes

Ngaire Underhill[1]  and Andrew Weinert[2]
MIT Lincoln Laboratory, Lexington, MA, 02420-9167, USA

## I. Introduction

The National Airspace System (NAS) is a complex and evolving system that enables safe and efficient aviation. Advanced air mobility concepts and new airspace entrants, such as unmanned aircraft, must integrate into the NAS without degrading overall safety or efficiency. For instance, regulations, standards, and systems are required to mitigate the risk of a midair collision between aircraft. Monte Carlo simulations have been a foundational capability for decades to develop, assess, and certify aircraft conflict avoidance systems. These are often validated through human-in-the-loop experiments and flight testing. For example, an update to the Traffic Collision Avoidance System (TCAS) mandated for manned aircraft was validated in part using this approach[1].

For many aviation safety studies, manned aircraft behavior is represented using the MIT Lincoln Laboratory statistical encounter models[2]–[5]. The original models[2]–[4] were developed from 2008-2013 to support safety simulations for altitudes above 500 feet Above Ground Level (AGL). However, these models were not sufficient to assess the safety of smaller unmanned aerial systems (UAS) operations below 500 feet AGL and fully support the ASTM F38 and RTCA SC-147 standards efforts. In response, newer models [5]–[7] with altitude floors below 500 feet AGL have been in development since 2018. Many of the models assume that aircraft behavior is uncorrelated and not dependent on air traffic services or nearby aircraft. The models were trained using observations of cooperative aircraft equipped with transponders but data sources and assumptions vary. The newer models are organized by aircraft types of fixed-wing multi-engine, fixed-wing single engine, and rotorcraft; while the original models do not consider aircraft type.

Our research objective was to compare the various uncorrelated models of conventional aircraft and identify how the models differ. Particularly if models of rotorcraft were sufficiently different than models of fixed-wing aircraft to require type specific models. The scope of this work was limited to altitudes below 5000 feet AGL, the expected

Distribution statement A: approved for public release. This material is based upon work supported by the Federal Aviation Administration under Air Force Contract No. FA8702-15-D-0001. Any opinions, findings, conclusions or recommendations expressed in this material are those of the author(s) and do not necessarily reflect the views of the Federal Aviation Administration. Delivered to the U.S. Government with Unlimited Rights, as defined in DFARS Part 252.227-7013 or 7014 (Feb 2014). Notwithstanding any copyright notice, U.S. Government rights in this work are defined by DFARS 252.227-7013 or DFARS 252.227-7014 as detailed above. Use of this work other than as specifically authorized by the U.S. Government may violate any copyrights that exist in this work. This document is derived from work done for the FAA (and possibly others), it is not the direct product of work done for the FAA. The information provided herein may include content supplied by third parties. Although the data and information contained herein has been produced or processed from sources believed to be reliable, the Federal Aviation Administration makes no warranty, expressed or implied, regarding the accuracy, adequacy, completeness, legality, reliability, or usefulness of any information, conclusions or recommendations provided herein. Distribution of the information contained herein does not constitute an endorsement or warranty of the data or information provided herein by the Federal Aviation Administration or the U.S. Department of Transportation. Neither the Federal Aviation Administration nor the U.S. Department of Transportation shall be held liable for any improper or incorrect use of the information contained herein and assumes no responsibility for anyone's use of the information. The Federal Aviation Administration and U.S. Department of Transportation shall not be liable for any claim for any loss, harm, or other damages arising from access to or use of data information, including without limitation any direct, indirect, incidental, exemplary, special or consequential damages, even if advised of the possibility of such damages. The Federal Aviation Administration shall not be liable for any decision made or action taken, in reliance on the information contained herein.
[1] Associate Staff, Surveillance Systems, AIAA Member.
[2] Associate Staff, Surveillance Systems, AIAA Senior Member, Corresponding Author, andrew.weinert@ll.mit.edu.

altitude ceiling for many new airspace entrants. The scope was also informed by the Federal Aviation Administration (FAA) UAS Integration Office and Alliance for System Safety of UAS through Research Excellence (ASSURE). The primary contribution is guidance on which uncorrelated models to leverage when evaluating the performance of a collision avoidance system designed for low altitude operations, such as prescribed by the ASTM F3442 detect and avoid standard for smaller UAS[8]. We also address which models can be surrogates for noncooperative aircraft without transponders. All models and software used are publicly available under open source licenses[9].

## II. Overview of the Encounter Models

Each encounter model is a set of two Bayesian Networks, generative statistical models that mathematically represent aircraft behavior during close or safety critical encounters, such as near midair collisions. The initial static Bayesian network represents the aircraft's initial position and dynamic state; while the transition network is a dynamic Bayesian network that represents state changes over time. In general, the initial network models initial aircraft position and kinematic state; while the transition network models subsequent aircraft flight in statistically realistic ways.

The original uncorrelated models from 2008-2013 were trained using observations collected by ground-based secondary surveillance radars from the 84th Radar Evaluation Squadron (RADES) network[2]–[4]; while recent models have been based on crowdsourced observations of Automatic Dependent Surveillance - Broadcast (ADS-B) equipped aircraft by the OpenSky Network[5]–[7]. Other models have also been trained based on different assumptions or data[10]–[13], but are not discussed here.

RADES data is filtered using the Mode 3A/C code squawked by an aircraft's transponder, which includes a non-unique identification code and a discretized MSL altitude. A Mode 3A code of 1200 indicates that the aircraft is operating under visual flight rules. In addition to Mode 3A/C, OpenSky Network ADS-B data also includes an ICAO 24-bit address, a unique global identifier[14]. This address is correlated with national aircraft registries to identify aircraft type (e.g. fixed-wing multi-engine, rotorcraft, etc.)[6]. The OpenSky Network-based models are organized based on aircraft type, while the RADES-based model is trained using multiple aircraft types. Each aircraft type was organized into two models, the first trained with tracks associated with aircraft squawking 1200 only and the second trained with tracks not squawking 1200.

The quantity and quality of training data is dependent on the coverage and characteristics of the surveillance system. Particularly, the low altitude coverage of the RADES and OpenSky networks widely varies across periods of time and location. The encounter models do not provide information on the density or frequency of aircraft behavior per unit time or a specific geographic region; other technologies address this[15].

Additionally, due to position uncertainties associated with ground-based radar time of arrival measurements, the RADES-based model has an altitude floor of 500 feet AGL. Conversely, ADS-B enables aircraft to broadcast the aircraft's estimate of their own location, which is often based on precise GNSS measurements, enabling OpenSky Network-based models to have an altitude floor of 50 feet AGL.

Table 1 summarizes the different uncorrelated models. With the exception of the lowest altitude bin, the uncorrelated models have the same structure and variable discretization. The model variables are geographic domain, airspace class, altitude layer, velocity, acceleration, vertical rate, and turn rate. The first three variables model aircraft location. Geographic domain denotes operations over the USA mainland, islands, or offshore environments. Airspace class denotes operations in the FAA Class B, C, D regulatory airspaces, or "other" airspace consisting of Class E and G airspaces; the airspace rules on aircraft behavior varies by class. The other variables correspond to assumed modeled aircraft kinematics. Refer to previous documentation[2]–[5] for details.

**Table 1: List of analyzed encounter models.**

| Aircraft Type | Source | Squawk Criteria | Altitude Floor |
|---|---|---|---|
| Unknown | RADES | 1200-code only | 500 feet AGL |
| Fixed-Wing Multi-Engine | OpenSky Network | 1200-code only | 50 feet AGL |
| Fixed-Wing Single Engine | OpenSky Network | 1200-code only | 50 feet AGL |
| Rotorcraft | OpenSky Network | 1200-code only | 50 feet AGL |
| Fixed-Wing Multi-Engine | OpenSky Network | 1200-code excluded | 50 feet AGL |
| Fixed-Wing Single Engine | OpenSky Network | 1200-code excluded | 50 feet AGL |
| Rotorcraft | OpenSky Network | 1200-code excluded | 50 feet AGL |

## III. Analysis

The analysis consisted of four components. First, a brief literature review focused on rotorcraft, as previous research focused on reviews of fixed-wing aircraft. Then, a comparison of the distributions for each initial network variable to identify evident differences between models. The next two analyses were based on the sampled joint distributions of pairs of initial network variables and distributions of kinematic variables sampled from the transition network. For these two analyses, one million trajectories were generated from each model using basic rejection sampling based on altitude and airspeed[4], [9, p.]. When sampling a given bin, we assume a uniform distribution within the bin. Each trajectory was 180 seconds long with one second updates.

### A. Literature Review

While rotorcraft have unique capabilities, such as the ability to hover, it is important to characterize how often these unique capabilities are used. As part of a pilot survey prior to prototyping a collision avoidance system on helicopters, Taylor and Adams[16] asserted that for most phases of flight, the flight profiles between a helicopter and general aviation aircraft were indistinguishable. This was due to passenger comfort and standard turn and vertical maneuvering rates. When testing a collision avoidance prototype equipped on helicopters, Harman et al.[17] and J.W. Andrews[18] discussed civil helicopters and fixed-wing aircraft have different fuselage shapes and that helicopters generally fly slower with greater turn rates, lower to the ground, and closer to buildings. They did not assert that the flight dynamics between the aircraft types were significantly different. Matthews and Sawyer[19] noted that for terminal operations under Instrument Flight Rules (IFR), rotorcraft tend to have slower speeds and greater turn rates than fixed-wing aircraft. Due to numerous surveillance, algorithm, and policy challenges, collision avoidance systems for rotorcraft were not mandated in the 1990s. Industry-wide research into collision avoidance systems for rotorcraft subsequently waned until recently.

### B. Initial Networks

First, we identified evident differences between some initial networks. Due to similar coverage bounds, the geographic domain distribution was extremely similar across all models. The distributions for the kinematic variables (acceleration, vertical rate, turn rate) were all indicative of a preference for straight and level flight. We assessed individually the models in two groups, the first group for 1200-code only models and the second for 1200-code excluded; with the results provided in figures 1 and 2 respectfully. An inspection of airspace class, altitude, and speed however revealed some significant model differences in Fig. 1-2. We emphasize that the model assumes inter-bin uniform distributions. For example, when training the model, two altitude observations of 600 and 1000 feet AGL, would both be discretized to [500, 1200) feet AGL. When sampling that bin, both values should equally be likely.

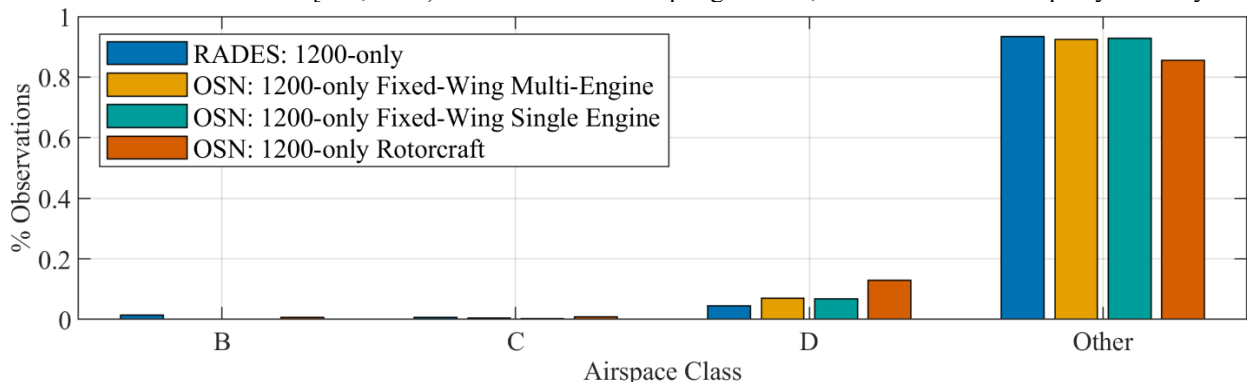

**a) Airspace class.**

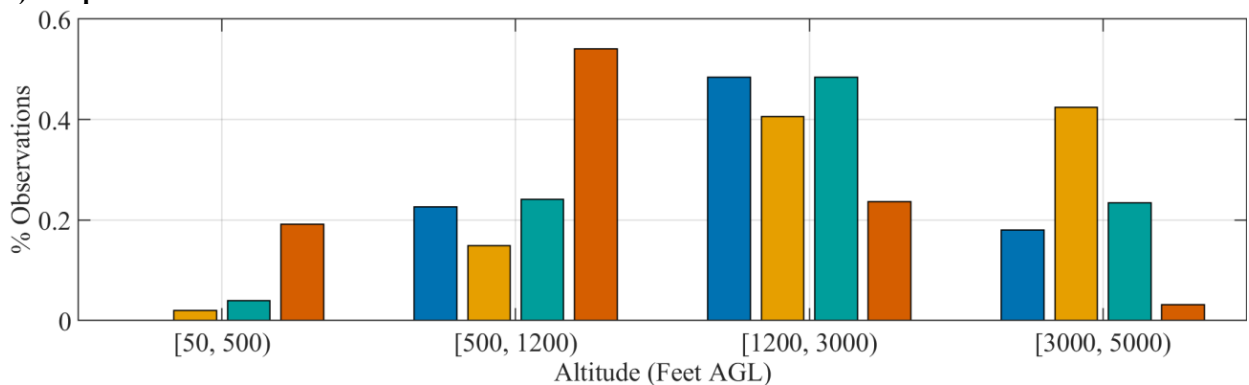

**b) Altitude layer.**

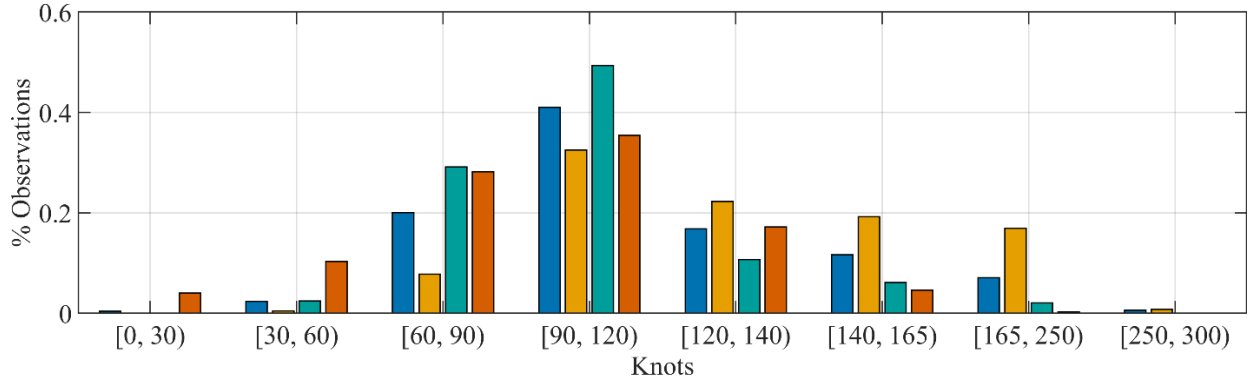

**c) Velocity.**

**Fig 1: Airspace, altitude, and velocity initial network distributions of 1200-code only encounter models.**

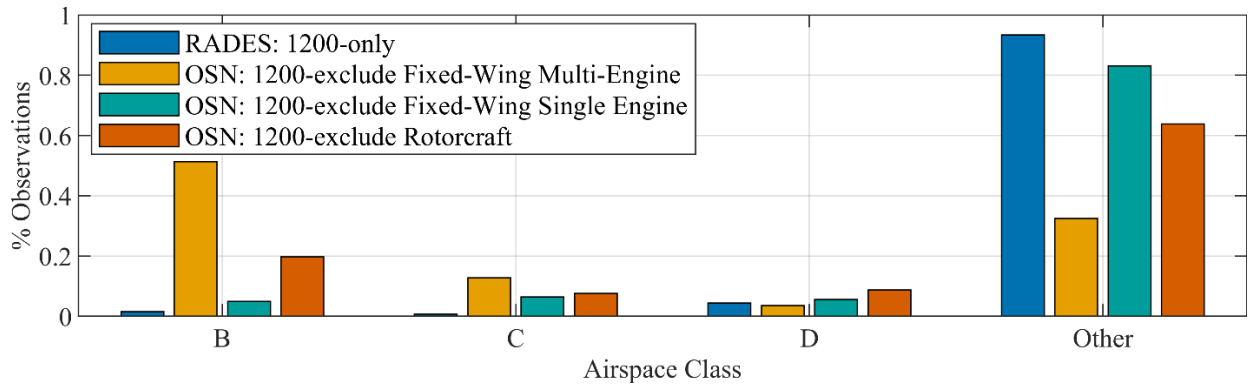

**a) Airspace class.**

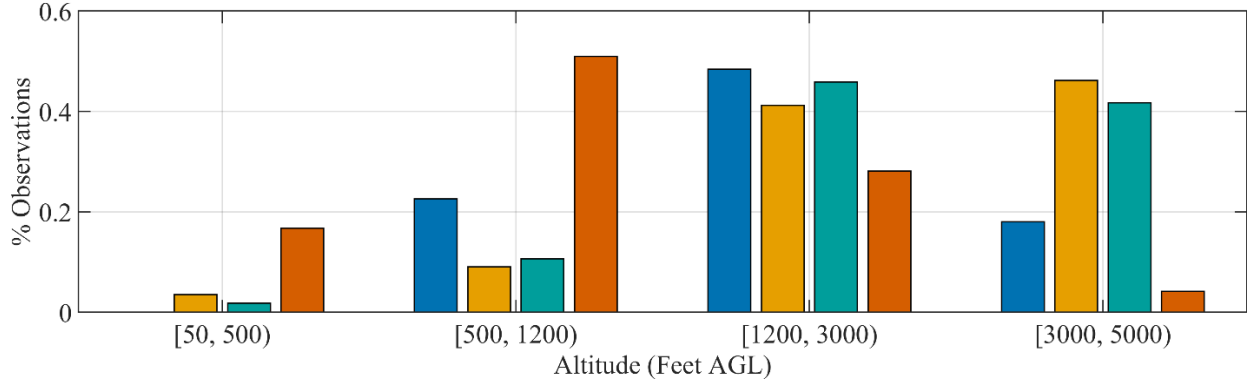

**b) Altitude layer.**

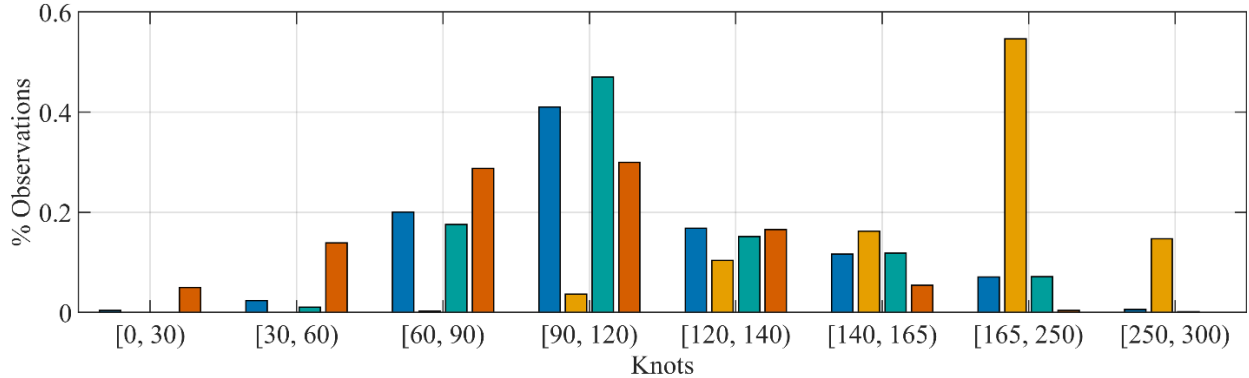

**c) Velocity.**

**Fig 2: Airspace, altitude, and velocity initial network distributions of 1200-code excluded encounter models.**

The initial networks aligned with many of the historical assumptions. Independent of the Mode 3/AC code, rotorcraft generally fly slower and lower. There were also limited modeled instances of rotorcraft flying at 30 knots or less. The 1200-code only and 1200-code excluded rotorcraft models were very similar. Both fixed-wing multi-engine models skewed towards higher altitudes and faster airspaces; while sharing limited similarities with the other models. The airspace class distribution was dependent on Mode 3/AC code, with a greater relative percentage of observations in terminal airspace for the 1200-code excluded models.

Conversely, there were significant differences between the 1200-code only and 1200-code excluded fixed-wing multi-engine models; with the latter trained with substantially more observations in Class B airspace and speeds of 165 knots or greater. Compared to other aircraft types, fixed-wing multi-engine aircraft were also more likely to accelerate at a magnitude greater than 0.3 knots per seconds. This isn't surprising, as large transport aircraft have multiple engines and squawk discrete code.

Also note the similarities between the 1200-code only RADES and fixed-wing single engine models. For all potential pairs of models, the distribution shapes and magnitudes were the most similar. These models differed from the 1200-code excluded fixed-wing single engine model, which had an altitude distribution skewed towards higher altitudes with slightly faster velocities

**C. Pairwise Overlap**

The previous analysis assessed only the independent variable distributions. Although informative for high-level trends, it was insufficient to characterize the models as it did not account for the Bayesian relationship between variables. This analysis considers pairs of variables. We calculated model overlap as the frequency that trajectories generated from different models were sampled from analogous or the same variable bins. An overlap of 85% and greater is indicative of significant similarities between the sampled trajectories, whereas 50% or less is indicative of poor overlap. We heuristically selected these thresholds, with 85% assumed to be significant overlap although accounting for minor model differences. Table 2 reports the overlap for notable variable combinations between the 1200-code only RADES and rotorcraft models. All variable pairs were assessed but for conciseness, only notable combinations are discussed here.

For example, Table 2 reports that 50% of sampled trajectories from the 1200-code only RADES model and both rotorcraft models were initialized in the same airspace class and same altitude layer bin. The lack of overlap was expected given the different altitude distributions in Fig 1. More than 50% of rotorcraft observations were at [500, 1200) feet AGL, compared to less than 25% for all other models. As velocity is dependent upon altitude, a lack of an overlap is observed again at 50%. However, there is better overlap when just considering the kinematics, with overlaps often greater than 85%. These results also align with the historical assumptions that rotorcraft fly similar to fixed-wing aircraft.

**Table 2: Overlap between 1200-code only RADES and rotorcraft sampled trajectories.**

| Dependent Variable | Independent Variable | Overlap Percentage | Note |
|---|---|---|---|
| Altitude | Geographic Domain | 90% | Location |
| Altitude | Airspace Class | 50% | Location |
| Velocity | Airspace Class | 76% | Location / Kinematic |
| Acceleration | Airspace Class | 93% | Location / Kinematic |
| Vertical Rate | Airspace Class | 90% | Location / Kinematic |
| Turn Rate | Airspace Class | 90% | Location / Kinematic |
| Velocity | Altitude | 50% | Location / Kinematic |
| Acceleration | Velocity | 78% | Kinematics |
| Vertical Rate | Acceleration | 88% | Kinematics |
| Turn Rate | Acceleration | 86% | Kinematics |
| Turn Rate | Vertical Rate | 93% | Kinematics |

Likewise, trends from the previous analysis were also identified here. Fixed-wing multi-engine trajectories had little to no overlap with the RADES trajectories; with majority of pairwise combinations having less than 50% overlap. Conversely, there was consistent overlap between the fixed-wing single engine and RADES-based trajectories. The 1200-code only fixed-wing single engine model has very good overlap, but even when blending the two fixed-wing single-engine models together, the minimum overlap with the RADES trajectories was 74% with many combinations greater than 85%. Comparing the 1200-code excluded rotorcraft trajectories to the RADES or fixed-wing single engine trajectories also drew a comparable conclusion of better overlap for kinematics than location.

### D. Transition Network

The previous analysis focused on aggregate trends on how trajectories were initialized. It did not characterize how an aircraft behaved over a complete trajectory as the transition network was sampled each timestep to update acceleration, vertical rate, and turn rate. We calculated the overlap between trajectories again and drew similar conclusions. For 1200-code only RADES and rotorcraft sampled trajectories, there was 88%, 93%, and 95% overlap for acceleration, vertical rate, and turn rate. Compared against all models, multi-engine aircraft had greater accelerations and vertical rates more frequently. All models had greater than 90% overlap for turn rate. This was likely due to pilots consistently employing standard rate turns.

Fixed-wing single engine and the RADES trajectories were again consistently the most similar. Fittingly, a blended model of fixed-wing single engine had overlaps of 90%, 98% and 99%; with both models independently also having significant overlap.

Interestingly even at slow airspeeds of 60 knots or less, there was significant overlap of greater than 90% for acceleration, vertical rate, and turn rate when comparing fixed-wing single engine and rotorcraft. This strongly supports the historical assumption that while rotorcraft can fly slower and lower than fixed-wing single engine, the underlying kinematics are similar.

## IV. Guidance for Model Use

An inspection of both the initial and transition network differences indicated that the models are representative of a variety of flight operations. The relative distribution of *where* aircraft were instantiated, as represented by geographic domain, airspace class, and altitude, differed between rotorcraft and fixed-wing single engine. Notably, rotorcraft were more likely to be instantiated at lower altitudes. However, the physical dynamics representing how aircraft fly, as modeled by speed, acceleration, vertical rate, and turn-rate were similar across models; with stronger similarities at faster speeds. Furthermore, while aircraft may maneuver in a similar manner, the frequency and magnitude of maneuvers may differ.

### A. Surrogacy for Noncooperative Aircraft

For conventional fixed-wing aircraft, the 1200-code only RADES model was previously determined to be surrogate for many noncooperative aircraft not equipped with any transponder [3]. This assumption was based on comparing the statistical model with a classifier trained using primary radar tracks with no transponder information. An identical analysis was not possible when training the OpenSky Network-based models, as the OpenSky Network due to concerns about using a classifier trained on one data source and applying it to another data source.

Across all analyses, there were consistent similarities between of the 1200-code only model trained using RADES data and fixed-wing single engine aircraft models trained using observations from the OpenSky Network. This strongly suggests that the behavior of fixed-wing aircraft is not dependent on whether the aircraft was equipped with ADS-B and observed by the OpenSky Network or if the aircraft was observed by a ground-based RADES radar. Given these similarities, we assume that the 1200-code only fixed-wing single engine model, trained with observations from the OpenSky Network, can represent simulated noncooperative fixed-wing single engine aircraft.

There was also strong evidence that fixed-wing single and rotorcraft aircraft employ similar, but not identical, accelerations, vertical rates, and turn rates. While the initial networks for rotorcraft are different, there is insufficient evidence that a rotorcraft squawking 1200 has significantly different kinematics than a fixed-wing single engine. Thus, we also assume that the 1200-code only rotorcraft model can represent noncooperative rotorcraft.

Lastly, while the initial networks of the 1200-code only RADES and fixed-wing multi engine models have similar shapes, the multi-engine models consistently generated tracks for aircraft that accelerated, climbed, and descended faster. These greater rates are not surprising when assuming the RADES model was trained primarily based on less powerful single engine aircraft. The analysis of initial networks also illustrated that fixed-wing multi-engine aircraft squawking 1200 behave significantly different than otherwise. Given a reasonable hypothesis for the kinematic differences between models and the behavioral dependence on squawking 1200, we assume that the 1200-code only fixed-wing multi-engine model can represent noncooperative fixed-wing multi-engine aircraft. However, this assumption is made with less confidence than for the other aircraft types and subject to future refinement.

### B. Applicability

We previously assumed the RADES model was trained primarily on observations of fixed-wing single engine with some minor mix of other aircraft types, but could not validate this assumption due to insufficient information provided by the Mode 3A/C transponders. Our analyses support this assumption because the 1200-code only model trained on RADES observations and the fixed-wing single engine models trained based on the OpenSky Network were consistently very similar. While these specific models were similar, we found that each of the aircraft type models trained using OpenSky Network data exhibited different behavior and are representative of particular types of traffic. Within the training data, rotorcraft were more likely to be flying lower and slower, while fixed-wing multi-engine were more likely to be flying higher and faster. There was significant evidence that fixed-wing aircraft, either single or multi-engine, behaved different if squawking 1200 code or not.

Accordingly, for safety simulations that require assessments for cooperative and non-cooperative aircraft the models are not interchangeable. For example, a simulation that only leverages the 1200-code only RADES model would not fully represent the low altitude environment and would likely be insufficient. Additionally, only simulating with both rotorcraft models would also be insufficient, as fixed-wing aircraft have a different distribution of speeds; the underlying kinematics would also be similar, but sufficiently different.

Thus, for safety simulations involving low altitude non-cooperative aircraft, only the 1200-code only models should be used. For simulating cooperative aircraft, the 1200-code excluded models should be used and then supplemented by the 1200-code only models. Specifically for evaluations related to ASTM F3442[8], all of the Open Sky Network-based aircraft specific models should be leveraged and use accordingly for assessing performance with cooperative and noncooperative manned aircraft. When sampling the models, at least the [50, 500) and [500, 1200) feet AGL altitude layers should be considered. An extensive use of all the models, along with uniform sampling within bins, supports a robust evaluation of if a system satisfies the ASTM F3442 or similar standard. For each unique model, the appropriate level of performance requirements should be satisfied. In the absence of traffic density and surveillance coverage information, it is difficult to justify weighting one model over another.

## V. Conclusion

The objective of this study to assess the similarities and differences between encounter models was satisfied. The assumptions that the models adequately represent rotorcraft and that fixed-wing single-engine flights enable safety simulations for altitudes below 500 feet AGL were validated. The similarities between 1200-code-only RADES and fixed-wing single-engine models also support that previous safety analyses were representative and valid.

This study determined that multiple models based on aircraft type are warranted and that model behavior can be dependent if an aircraft is squawking 1200-code. A consequence of this research is that multiple models should be used when assessing safety systems at lower altitudes. The recommendation for multiple models differs from assessments before 2018, where the singular RADES 1200-code model was solely leveraged for most simulations.

## Acknowledgments

We greatly appreciate the support and assistance provided by Sabrina Saunders-Hodge and Adam Hendrickson from the Federal Aviation Administration UAS Airspace Integration Office under Air Force Contract No. FA8702-15-D-0001. We also like to thank our colleagues Dr. Jim Kuchar and Dr. Rodney Cole.